\documentclass{article}

\usepackage{microtype}
\usepackage{graphicx}
\usepackage{subfigure}
\usepackage{booktabs} 

\usepackage{hyperref}

\usepackage[accepted]{icml2024}

\usepackage{amsmath}
\usepackage{amssymb}
\usepackage{mathtools}
\usepackage{amsthm}
\usepackage{interval}

\usepackage[capitalize,noabbrev]{cleveref}

\usepackage{float}

\theoremstyle{plain}

\theoremstyle{definition}

\theoremstyle{remark}

\usepackage[textsize=tiny]{todonotes}

\icmltitlerunning{Learned Best-Effort LLM Serving}

\begin{document}

\twocolumn[
\icmltitle{Learned Best-Effort LLM Serving}



\icmlsetsymbol{equal}{*}

\begin{icmlauthorlist}
\icmlauthor{Siddharth Jha}{berkeley}
\icmlauthor{Coleman Hooper}{berkeley}
\icmlauthor{Xiaoxuan Liu}{berkeley}
\icmlauthor{Sehoon Kim}{berkeley}
\icmlauthor{Kurt Keutzer}{berkeley}
\end{icmlauthorlist}

\icmlaffiliation{berkeley}{UC Berkeley}

\icmlcorrespondingauthor{Siddharth Jha}{sidjha@berkeley.edu}

\icmlkeywords{Machine Learning, ICML}

\vskip 0.3in
]



\printAffiliationsAndNotice{} 
\begin{abstract}
Many applications must provide low-latency LLM service to users or risk unacceptable user experience. However, over-provisioning resources to serve fluctuating request patterns is often prohibitively expensive. In this work, we present a best-effort serving system that employs deep reinforcement learning to adjust service quality based on the task distribution and system load. Our best-effort system can maintain availability with over $10\times$ higher client request rates, serves above $96\%$ of peak performance $4.1\times$ more often, and serves above $98\%$ of peak performance $2.3\times$ more often than static serving on unpredictable workloads. Our learned router is robust to shifts in both the arrival and task distribution. Compared to static serving, learned best-effort serving allows for cost-efficient serving through increased hardware utility. Additionally, we argue that learned best-effort LLM serving is applicable in wide variety of settings and provides application developers great flexibility to meet their specific needs.

\end{abstract}

\section{Introduction}
Applications in the last decade have evolved from using machine learning in the background for tasks such as data analytics and monitoring to now being at the forefront of user experience. Many applications are using large language models (LLMs) to provide users with both custom and interactive experiences through chat agents and virtual assistants. The need for latency guarantees is critical for such applications as applications cannot simply become unavailable and unresponsive to users. This presents a challenge for LLM serving, as a simple solution of over-provisioning GPU resources to run models in parallel in order to serve bursty request windows is prohibitively expensive for small businesses and independent application developers. Another strategy may be to use a smaller model that serves at lower latency. However, naively using a small model in place of a large model can lead to undesirable quality degradation.

\begin{figure}[t]
\vskip 0.1in
\begin{center}
\centerline{\includegraphics[width=\columnwidth]{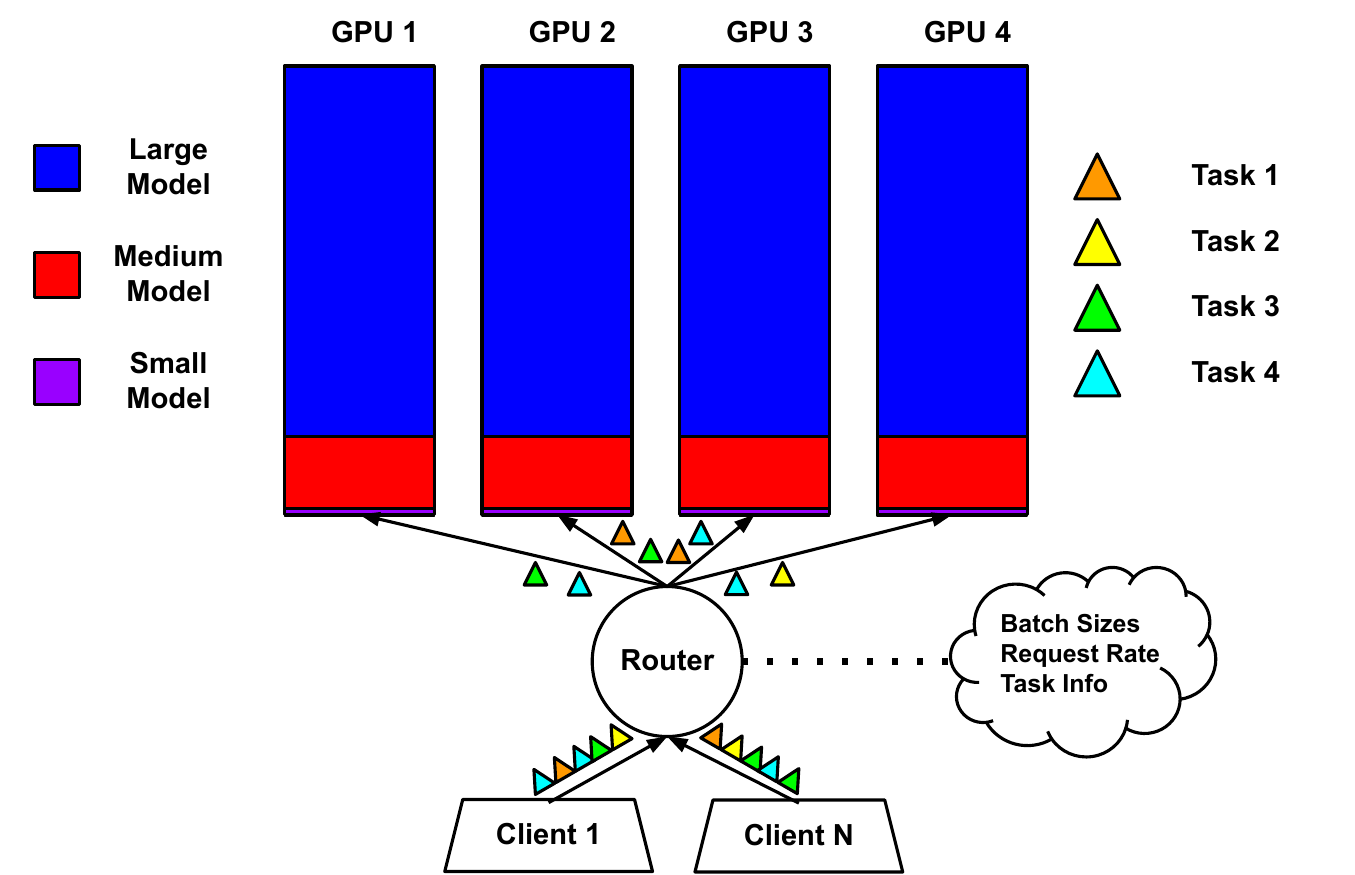}}
\caption{Learned best-effort serving consists of multiple models serving multiple tasks, with a router that keeps track of system load and task information in order to route requests to models. In this example, each model is replicated on 4 GPUs, but any model partitioning and replication scheme may be used. Additionally, any number of models and tasks may be used.}
    \label{fig:env-1}
\end{center}
\vskip -0.25in
\end{figure}

In response to the challenge of delivering low-latency LLM services without additional hardware costs, we introduce a learned best-effort serving framework that dynamically selects models of varying sizes to match client requests, guided by a model-routing mechanism as illustrated in \autoref{fig:env-1}. Leveraging the lower memory footprint of smaller models, best-effort serving does not inhibit the ability to run the largest desired model in the system, similar to speculative inference~\citep{leviathan2023fast} methods.

Our system is designed to handle requests across a spectrum of tasks, balancing between model size for accuracy and meeting deadlines, categorized as hard or soft, to optimize overall performance. Performance is quantified by accuracy for hard deadlines, and decreases as soft deadlines are exceeded. The ultimate objective is to maximize cumulative performance, with "peak performance" considered as the ideal state of meeting all deadlines using the largest model. Our general framework allows developers to adjust the balance between quality and latency. We demonstrate that effective request routing is influenced by the set of tasks, the task distribution from incoming requests, and system load. We employ deep reinforcement learning with minimal hyper-parameter tuning through the DQN algorithm to manage these dynamics efficiently.

In summary, learned best-effort serving provides a variety of benefits over traditional static serving methods for low-latency LLM applications:

\begin{itemize}
    \item \textbf{Performance:} It maintains over $96\%$ of peak performance $4.1\times$ more frequently, and exceeds $98\%$ of peak performance $2.3\times$ more frequently than static serving with a large model on unpredictable workloads.
    \item \textbf{Availability:} Learned best-effort serving can meet client deadlines at over $10\times$ higher system load than static serving with a large model. Compared to static serving with a medium sized model, learned best-effort serving achieves at least $94\%$ of peak performance $28.21\times$ more often while still providing higher availability.
    \item \textbf{Cost-Efficiency:} When compared to a static serving system that uses twice the GPUs, learned best-effort serving still surpasses $90\%$ of peak performance $1.51\times$ more often. Furthermore, learned best-effort serving delivers $3.94\times$ higher performance per hardware unit.
\end{itemize}

\section{Preliminaries}
\subsection{Efficient LLM Serving}
\label{sec:llms}
The use of multiple models to serve LLM requests has been explored via speculative inference~\citep{leviathan2023fast, spector2023accelerating}, which uses a small draft model to generate tokens to be verified by a large model. Big Little Decoder~\citep{kim2023big} is a speculative inference technique that allows clients to adjust quality and latency by changing hyper-parameters. However, this requires a hyper-parameter search for every task and does not adjust under load. Furthermore, serving speculative inference in real deployments with optimizations such as continuous batching~\citep{yu2022orca} is difficult. Dynamically adjusting serving in response to load has been explored though autoscalers. Autoscalers such as Ray~\citep{moritz2018ray} dynamically increase GPU instances under load. However, acquiring on-demand GPU instances is expensive and not instantaneous. Model switching has been explored in~\citep{zhang2020model, eccles2024dnnshifter}. However,~\citep{zhang2020model} only considers CNN models running on CPUs and does not consider the task when selecting a model. Furthermore,~\citep{eccles2024dnnshifter} necessitates pruned models and can only have one active model at a time.

\subsection{Deep Reinforcement Learning}
Deep RL is a promising technique for learning to control systems, and it has been successfully applied in a variety of areas such as continuous controls~\citep{brockman2016openai} and games~\citep{mnih2013playing}. There are three core components in any RL problem: states, actions, and rewards. The RL policy aims to maximize the total rewards it sees as it takes actions and transitions between states. Deep Q-learning methods learn a Q-function, represented as a neural network, that map state-action pairs to the expected return of taking the action in the state and then following the policy. After fitting the Q-function of the optimal policy, the Q-function may be used to select actions with the highest expected reward. Popular algorithms in this area include DQN~\citep{mnih2013playing}, Double Q-learning~\citep{van2016deep}, and PER~\citep{schaul2015prioritized}.

\section{Best-Effort Serving}
\subsection{Problem Formulation}
We envision a system where multiple clients dispatch requests for LLM inference. These requests are aligned with predefined tasks (e.g. summarization, question answering, etc.) tailored to the application, each tagged with a designated latency requirement. Service requirements are determined at the task granularity, meaning that latency requirements are equivalent for requests belonging to the same task. The system's infrastructure categorizes deadlines into two types: hard and soft. For hard deadlines, the utility of a response is gauged by its accuracy if it meets the request's latency requirement. Conversely, for soft deadlines, a response's utility diminishes proportionally with the extent of delay past the deadline. The overarching objective of the system is to optimize the cumulative utility across all client requests.

Suppose the best-effort system is serving $T$ tasks using $M$ model choices. Let $A\in\mathbb{R}^{T\times M}$ be a matrix so that $A_{tm}$ denotes the client utility (e.g. accuracy) of serving a request from task $t$ using model $m$. Each client request is tagged with a task and deadline. Given a sequence of requests $(r_{n})_{n\in\mathbb{N}}$, let request $r_{n}$'s task be denoted by a one-hot encoded vector $t_{n}\in\mathbb{R}^{1\times T}$ and its latency requirement be denoted as $d_{n}\in\mathbb{R}$. If we let $m_{n}\in\mathbb{R}^{1\times M}$ be a one-hot encoded vector representing the model assigned to request $r_{n}$, the goal of the serving system is to optimize

\begin{equation}\label{eq:eq-1}
\max_{m} \sum_{n\geq1} w_{n}t_{n}Am_{n}^{T} \quad \text{subject to} \quad \|m_{n}\|_1 = 1, \forall n
\end{equation}

where $w_{n}$ is dependent on the deadline being satisfied. For applications with hard deadlines, $w_{n}\in\{0, 1\}$ is binary. When using soft deadlines, $w_{n}\in\interval[]{0}{1}$ decreases as the assigned deadline is further violated. The decay function that governs this may be set by application developers to best meet their requirements. Optimizing \autoref{eq:eq-1} for online serving is difficult as the system does not have access to the behavior of future requests when making a model selection decision for a present request. Static serving methods assign the same model to each request. In contrast, best-effort serving employs dynamic routing to various models, aiming to leverage the latency-quality trade-off most effectively. This methodology seeks to enhance the system's ability to meet diverse client demands while maximizing overall performance.

\subsection{Dynamic Router Agent}
To address the challenge of optimizing \autoref{eq:eq-1} in online settings, we employ deep reinforcement learning to learn a router agent that dynamically dispatches client requests to models. Since the agent cannot have access to information about future request patterns when making a routing decision, it must rely only the present and past. Thus the agent keeps track of the present batch size at each model and an approximation of the current arrival rate of the client request process using past requests. In sum, the state that the agent conditions on when routing consists of the task (numbered 0 though $T - 1$), the batch size at each model, and the current arrival rate. The action space consists of $M$ options, with each action corresponding to a separate model. The reward for picking an action in a given state is the exact same as in \autoref{eq:eq-1}. Thus the router agent is optimizing a proxy of \autoref{eq:eq-1}. To learn an effective policy for our router agent, we utilize the DQN algorithm with Double Q-learning to prevent over-estimation of Q-values and represent the policy as a 2-layer MLP with hidden size 256. As we do not assume prior information about the production workload, we train the policy by randomly switching between random arrival rates with a uniform task distribution. It is possible for application developers to exploit knowledge about their workloads to change the training distribution to closely match deployment, but our experiments in \autoref{sec:evaluation} show that random training works well in practice.

As we target latency-sensitive online serving workloads, it is critical that the agent runs efficiently with minimal resource consumption. Since the policy is a small MLP, it can perform inference nearly instantly compared to LLM inference. Specifically, when running the policy on our system's CPU, its inference latency is one-tenth of a millisecond. In contrast, inference of one token with a small OPT-125M model on our system's GPUs takes five milliseconds when prompted with a short sequence at batch size one. We give further information about our system's hardware in \autoref{sec:appendix-training-details}.

\section{Evaluation}
\label{sec:evaluation}

In \autoref{sec:exp-setup}, we describe the training procedure, baselines, and evaluated workloads. In \autoref{sec:setup}, we describe the models and tasks present in the evaluated serving system. In \autoref{sec:stable-workload} and \autoref{sec:unpredictable-workloads}, we show that learned best-effort serving outperforms static baselines in both stable and unpredictable workloads, respectively. We show that this holds with both hard and soft deadlines. In \autoref{sec:robustness}, we show that the learned policy is robust to shifts in the task distribution. In \autoref{sec:hw-utility}, we compare learned best-effort serving against a system using twice the GPUs. We show that learned best-effort serving provides both better quality of service as well as significantly higher performance per hardware unit. In \autoref{sec:diff-deadlines}, we demonstrate the policy's ability to learn and outperform baselines with different deadlines for different requests. Overall, our evaluation highlights multiple advantages provided by learned best-effort serving in a variety of settings.

\subsection{Experiment Setup}
\label{sec:exp-setup}
We use one set of hyper-parameters, further detailed in \autoref{sec:appendix-training-details}, for all trained policies. As baselines, we evaluate against static serving with just one model size. When running the baselines, we give all the GPU memory to each model. Models are served with vLLM~\citep{kwon2023efficient}. Graphs with uncertainty regions represent one standard deviation over three trials.

Prior work on model serving~\citep{li2023alpaserve,zhang2023shepherd,gujarati2020serving} uses Microsoft's Azure Function (MAF) traces~\citep{shahrad2020serverless, zhang2021faster} to model behavior of clients in a serving system. The MAF1 trace~\citep{shahrad2020serverless} consists of stable and steady request periods. On the other hand, the MAF2 trace~\citep{zhang2021faster} has much more unpredictable client behavior and the arrival rates rapidly change. Based off of these observations, we evaluate our system on three types of synthetic workloads that capture a wide range of client behavior. One workload is stable, while the other two are unpredictable. The synthetic workloads are generated in similar ways to~\citep{yu2022orca, zhang2023shepherd}, which use Poisson processes and Markov-modulated Poisson processes. We list further details about the three synthetic workloads and arrival rate estimation in \autoref{sec:appendix-workload-details}.

\subsection{System Setup}
\label{sec:setup}

To evaluate our routing policy, we consider a serving system with 4 GPUs. Each GPU contains an instance of OPT-125M, OPT-1.3B, and OPT-6.7B and there are 4 tasks in the serving system. Thus the serving system is equivalent to the one shown in \autoref{fig:env-1}. Smaller models can fit in device memory with the large model because the memory required for both their model parameters, key-value cache, and activations is significantly smaller than those of the large model. Specifically, we give $5\%$ of GPU memory to OPT-125M, $20\%$ of GPU memory to OPT-1.3B, and the rest to OPT-6.7B. When the router chooses a model size for the request, we automatically load balance by sending to the replica with the smallest batch size for the model.
We set the latency guarantee to be 40 milliseconds/token. Additionally we use zero-shot HellaSwag~\citep{zellers2019hellaswag}, COPA~\citep{roemmele2011choice}, PIQA~\citep{Bisk2020}, and OpenBookQA~\citep{OpenBookQA2018} as the four tasks in the system. We use each model's average accuracy on each task as a measure of its quality. For each task we normalize the accuracy of each model to OPT-6.7B's accuracy to get the rewards shown in \autoref{tab:env-1-rewards}. We train the policy for 1.2 million iterations using hard deadlines.

\begin{table}[h]
\caption{Rewards for tasks served in the system.}
\label{tab:env-1-rewards}
\vskip 0.15in
\begin{center}
\begin{small}
\begin{sc}
\begin{tabular}{lccc}
\toprule
Task & OPT-125M & OPT-1.3B & OPT-6.7B \\
\midrule
HellaSwag   &0.45&0.78&1.00 \\
COPA        &0.80&0.95&1.00 \\
PIQA        &0.82&0.96&1.00 \\
OpenBookQA  &0.70&0.94&1.00 \\
\bottomrule
\end{tabular}
\end{sc}
\end{small}
\end{center}
\vskip -0.1in
\end{table}

\subsection{Stable Workload}
\label{sec:stable-workload}
\subsubsection{Hard Deadlines}
\begin{figure}[h]
\vskip 0.1in
\begin{center}
\centerline{\includegraphics[width=\columnwidth]{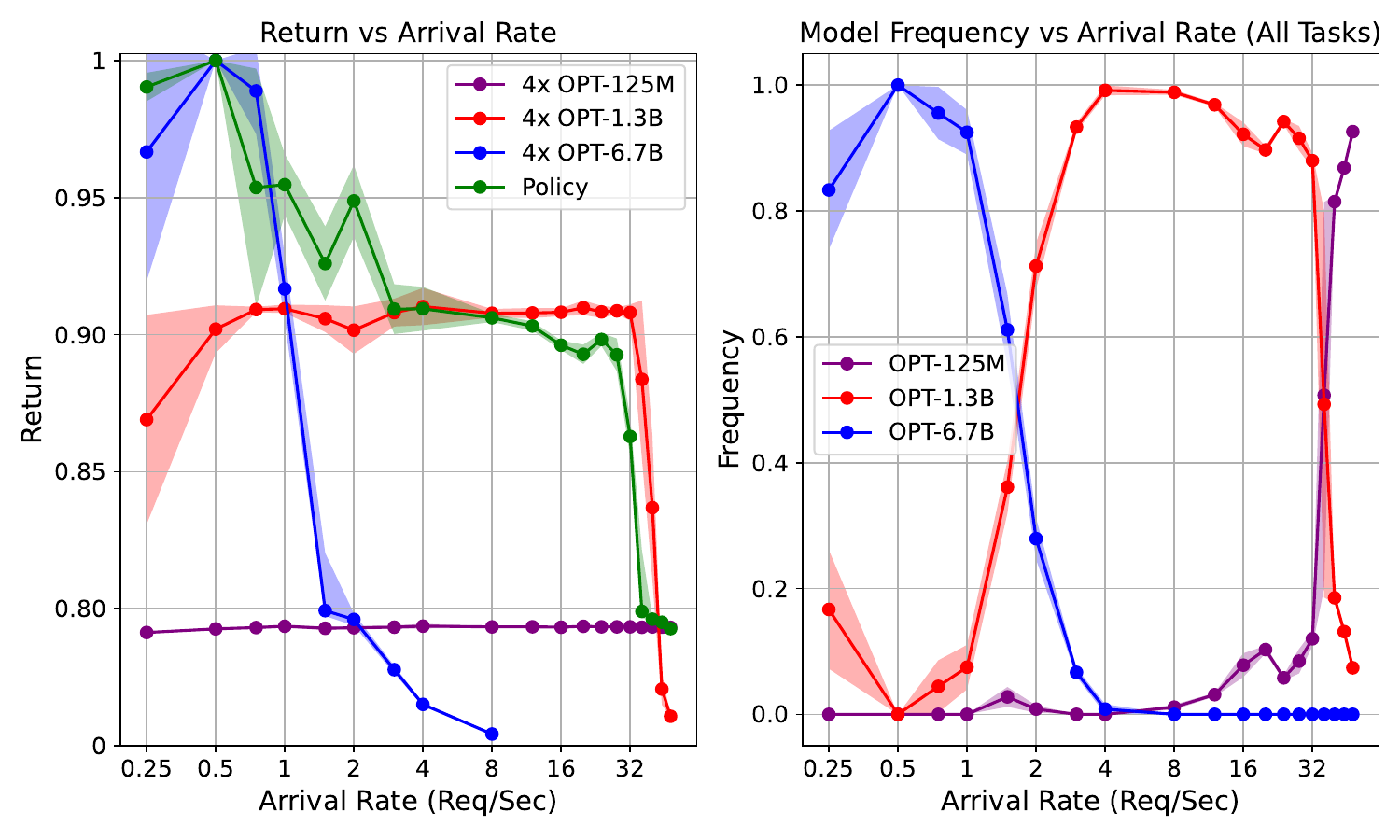}}
\caption{The left figure shows the performance with hard deadlines. The right figure shows the distribution of model selection from the policy.}
    \label{fig:env-1-hard-performance}
\end{center}
\vskip -0.25in
\end{figure}

For the stable workload, we vary the arrival rate of the arrival Poisson process from 0.25 to 48 requests per second and serve for 40 seconds at each arrival rate before resetting and going to the next arrival rate. We show the results with hard deadlines in \autoref{fig:env-1-hard-performance}. As baselines, we show the performance when only serving one of OPT-6.7B, OPT-1.3B, or OPT-125M.

\begin{figure}[h]
\vskip 0.1in
\begin{center}
\centerline{\includegraphics[width=\columnwidth]{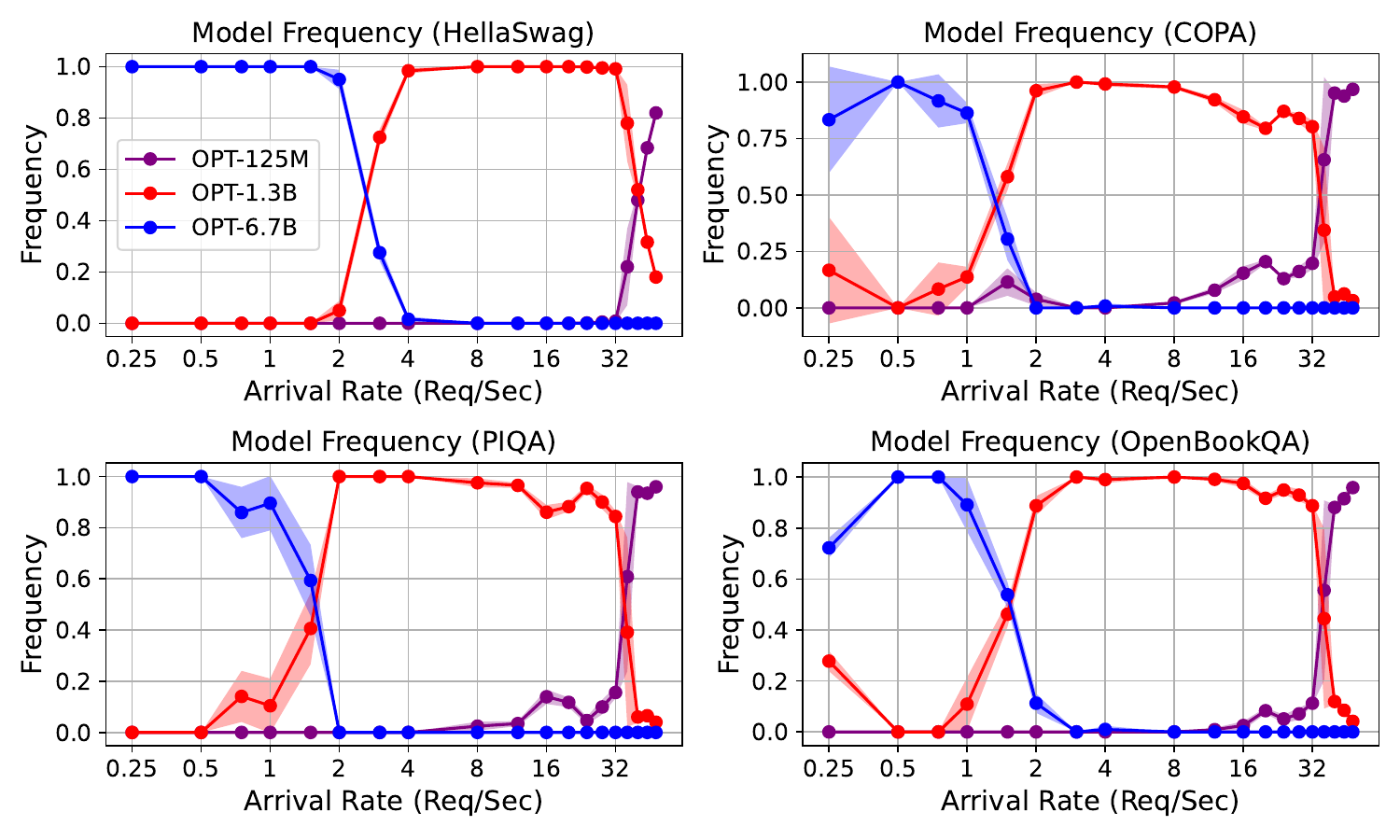}}
\caption{Model selection frequency for each individual task with hard deadlines.}
    \label{fig:env-1-hard-task-dist}
\end{center}
\vskip -0.25in
\end{figure}

As \autoref{fig:env-1-hard-performance} shows, in typical systems that serve all requests to OPT-6.7B, the performance is near the peak possible performance at low arrival rates. However, once the arrival rate increases past a threshold (2 requests per second), many latency deadlines are missed and performance sharply declines. While OPT-1.3B can serve requests at much higher arrival rates, its quality cannot match OPT-6.7B even when the arrival rate is low. Additionally, there is also a point at which OPT-1.3B cannot keep up with client requests. Serving only with OPT-125M leads to significant performance degradation at all but extremely high arrival rates.

In contrast, the policy dynamically adjusts which model to send requests to. When the arrival rate is low, the policy primarily sends to OPT-6.7B and achieves similar performance. However, as the arrival rate increases, the policy correctly routes more requests to OPT-1.3B and eventually even OPT-125M at the extreme end. Therefore the policy allows the system to remain available for over 10x faster arrival rates than just using OPT-6.7B while still providing equal quality to OPT-6.7B at low arrival rates. Furthermore we notice that there are regions where the policy even performs better than just taking the maximum of each of the baseline's curves in \autoref{fig:env-1-hard-performance} as it is able to multiplex between models at a given arrival rate.

We now examine how the routing varies for different tasks, as shown in \autoref{fig:env-1-hard-task-dist}. We see that the policy sends HellaSwag requests to OPT-6.7B much more often than the other three tasks. Taking a look at \autoref{tab:env-1-rewards}, we see that OPT-125M and OPT-1.3B have a significant quality gap compared to OPT-6.7B for HellaSwag. This quality gap is much larger than the gap between models on COPA, PIQA, and OpenBookQA. Therefore the policy appropriately learns to prioritize sending HellaSwag to the large model when possible. Furthermore, even when the arrival rate is higher, HellaSwag is sent to OPT-1.3B more often than the other three tasks, which are more frequently sent to OPT-125M. Thus the router learns a complex relationship not only depending on the task's quality across models in isolation, but with respect to the quality of other tasks in the system and their distribution.

\subsubsection{Soft Deadlines}
\begin{figure}[h]
\vskip 0.1in
\begin{center}
\centerline{\includegraphics[width=\columnwidth]{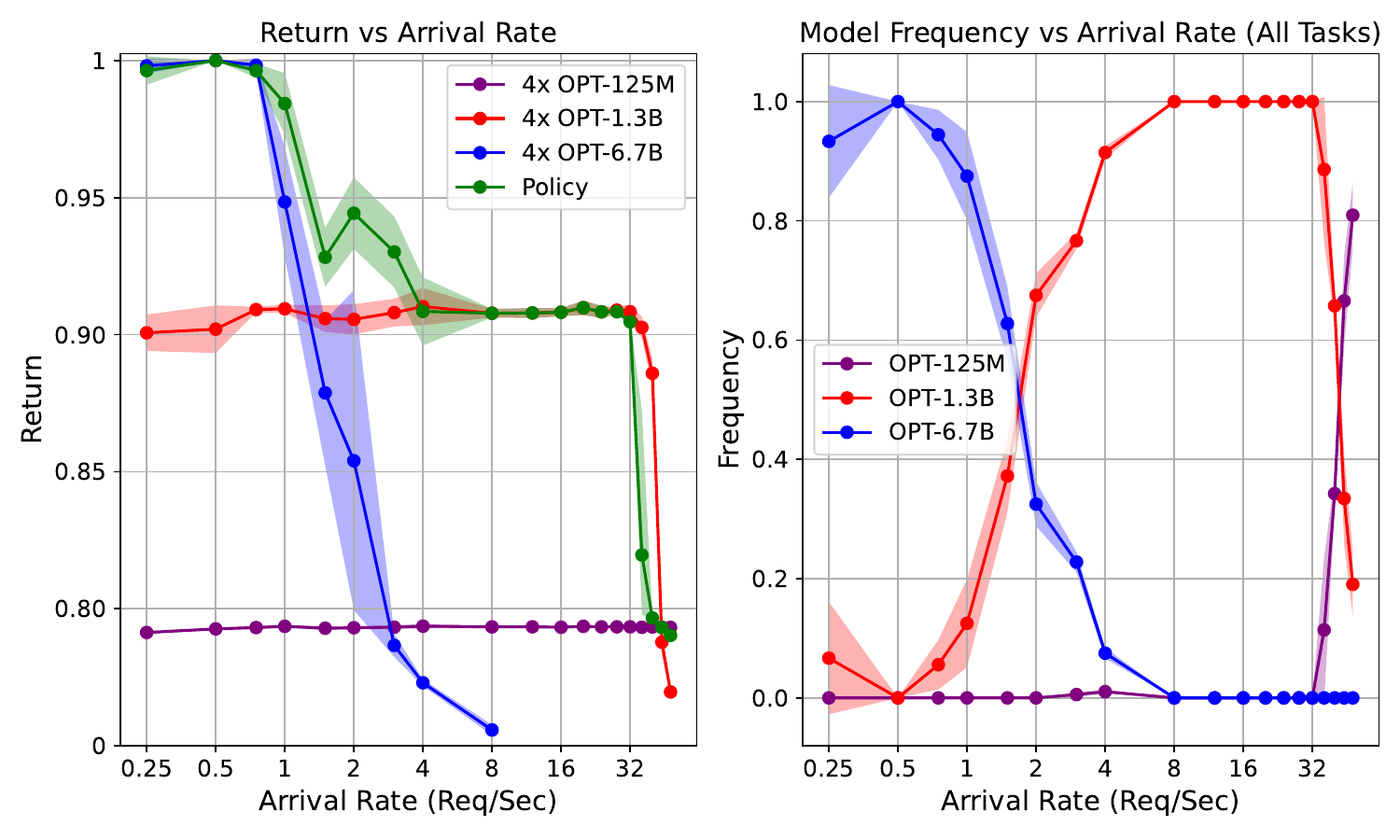}}
\caption{The left figure shows the performance with soft deadlines. The right figure shows the distribution of model selection from the policy.}
    \label{fig:env-1-soft-performance}
\end{center}
\vskip -0.25in
\end{figure}

We pick a specific soft deadline decay function and fine-tune the policy from the policy trained with the hard deadlines. It is also possible to train the soft policy from scratch. When fine-tuning, we adjust the reward function to be soft and train for an additional 685,000 iterations. Specifically, this soft deadline decays the reward by 1\% for every millisecond past the deadline as long as the violation is less than 10\% of the specified deadline. However, once the acceptable latency is violated by more than 10\%, the client does not value the response and the reward is zero. We show the results in \autoref{fig:env-1-soft-performance}. We see that the policy outperforms the baselines and sends more requests to larger models when using this soft deadline than when using hard deadlines. Compared to the hard policy's performance in \autoref{fig:env-1-hard-performance}, we see that the soft policy more closely follows OPT-1.3B's performance before switching to OPT-125M's performance. With hard deadlines, the policy takes a slightly more conservative approach in this regime and sends a small set of requests to the small model in order to prevent missing deadlines, creating a small performance gap between the policy and OPT-1.3B before the policy switches to OPT-125M's performance. With soft deadlines, the policy is less conservative and is able to almost exactly match the performance of OPT-1.3B at these arrival rates.

\begin{figure}[h]
\vskip 0.1in
\begin{center}
\centerline{\includegraphics[width=\columnwidth]{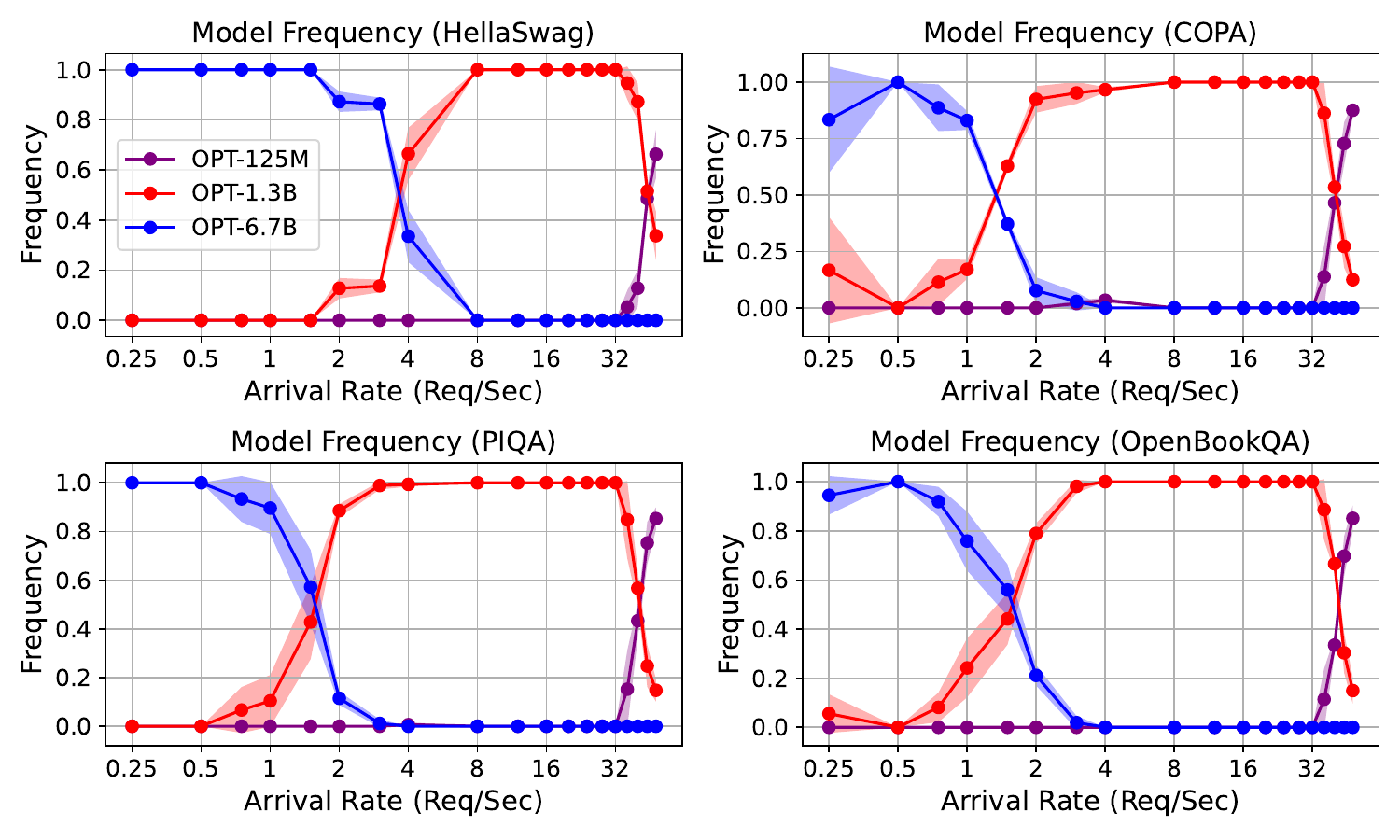}}
\caption{Model selection frequency for each individual task with soft deadlines.}
    \label{fig:env-1-soft-task-dist}
\end{center}
\vskip -0.25in
\end{figure}

We show how tasks are routed to models when using this soft deadline in \autoref{fig:env-1-soft-task-dist} and observe similar trends to \autoref{fig:env-1-hard-task-dist}. When measuring the usage of OPT-6.7B via a Riemann sum of the selection distribution, we see that HellaSwag's OPT-6.7B usage increases by $52\%$ with soft deadlines compared to hard deadlines. In contrast, PIQA, COPA, and OpenBookQA's OPT-6.7B usage increases by just $9\%$, $7\%$, and $3\%$, respectively. Thus the policy is able to exploit the leniency given by the soft deadline to reap the large gains in quality by sending HellaSwag to OPT-6.7B instead of OPT-1.3B.

\begin{table}[h]
\caption{Approximation of OPT-6.7B usage across tasks for both hard and soft deadlines, as measured by Riemann sum of selection distribution.}
\label{tab:env-1-soft-hard-area}
\vskip 0.15in
\begin{center}
\begin{small}
\begin{sc}
\begin{tabular}{lccc}
\toprule
Task & Hard & Soft & Percent Change \\
\midrule
HellaSwag   &2.53&3.86&$+52\%$ \\
COPA        &1.08&1.16&$+7\%$ \\
PIQA        &1.22&1.33&$+9\%$ \\
OpenBookQA  &1.30&1.34&$+3\%$ \\
\bottomrule
\end{tabular}
\end{sc}
\end{small}
\end{center}
\vskip -0.1in
\end{table}

\subsection{Unpredictable Workloads}
\label{sec:unpredictable-workloads}
\label{sec:env-1-unpredictable}
\begin{figure}[h]
\vskip 0.1in
\begin{center}
\centerline{\includegraphics[width=\columnwidth]{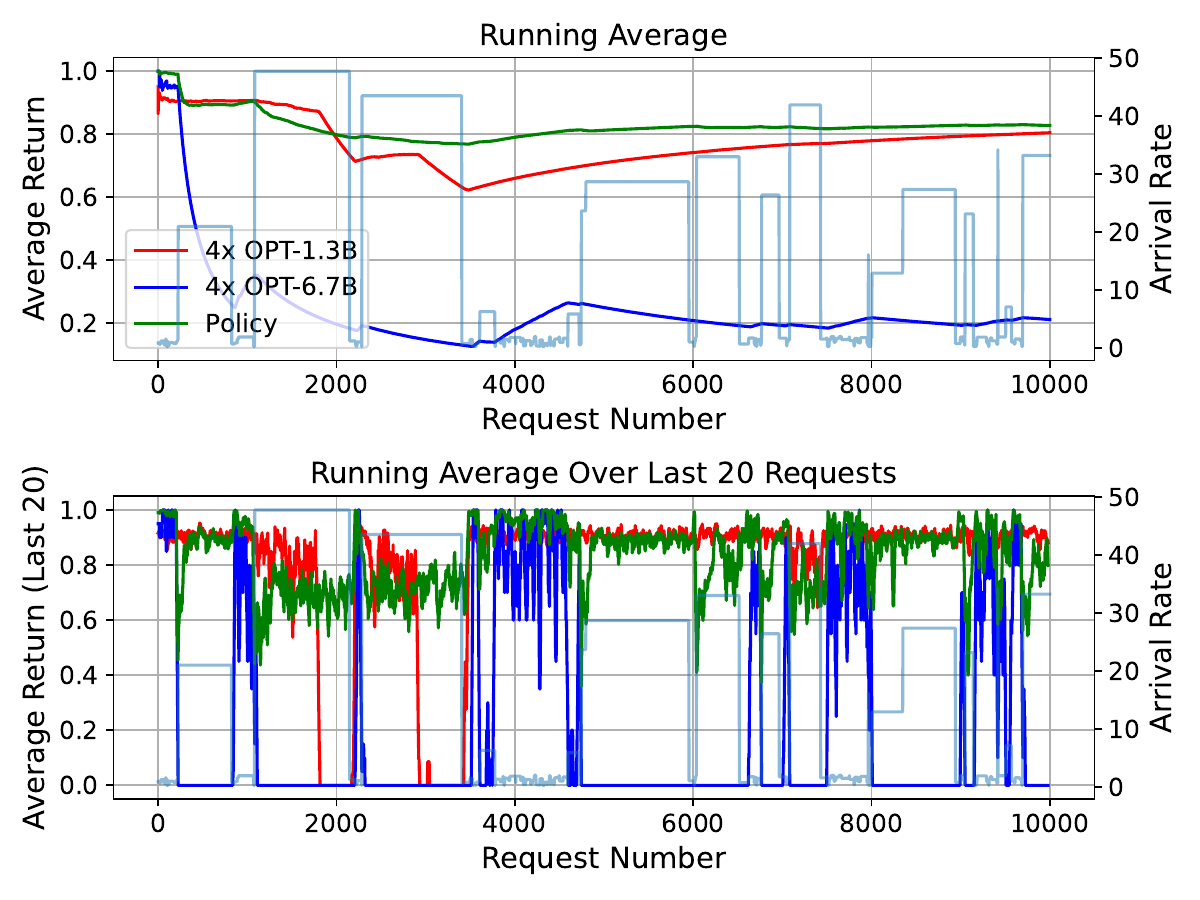}}
\caption{Running total and windowed average over the last 20 requests of performance on the first unpredictable workload. The arrival rate at each step is also shown.}
    \label{fig:env-1-unpredictable-1}
\end{center}
\vskip -0.25in
\end{figure}

We evaluate on two unpredictable workloads using hard deadlines with large bursts. Both workloads have different arrival patterns than the training workload. \autoref{fig:env-1-unpredictable-1} shows the performance of the routing policy as well as the baselines, in addition to the changing arrival rate for the first unpredictable workload. We show both the running average of performance across all served requests and the running average of the performance across the last 20 requests. The serving system that only uses OPT-6.7B fails to meet latency deadlines during many of the bursts and thus its performance is highly variable. Even though OPT-6.7B has more windowed averages at peak performance, the policy is able to perform at near-peak performance significantly more often. We quantify this in \autoref{tab:env-1-unpredictable-1-peak-nums}.

\begin{table}[h]
\caption{Number of request windows of size 20 that meet average performance thresholds on the first unpredictable workload.}
\label{tab:env-1-unpredictable-1-peak-nums}
\vskip 0.15in
\begin{center}
\begin{small}
\begin{sc}
\begin{tabular}{lccc}
\toprule
Threshold & Policy & OPT-6.7B & OPT-1.3B \\
\midrule
$= 1.00$    &142&\textbf{307}&0\\
$\geq 0.99$    &\textbf{470}&307&0\\
$\geq 0.98$    &\textbf{713}&307&0\\
$\geq 0.96$    &\textbf{1264}&307&0\\
$\geq 0.94$    &\textbf{1723}&625&154 \\
\bottomrule
\end{tabular}
\end{sc}
\end{small}
\end{center}
\vskip -0.1in
\end{table}

As shown in \autoref{tab:env-1-unpredictable-1-peak-nums}, Compared to the policy, OPT-6.7B is able to achieve more windowed averages with peak performance. However, when analyzing the number of windows which meet high performance thresholds such as 0.99, 0.98, 0.96, and 0.94, the policy achieves more such windows than OPT-6.7B and OPT-1.3B. For example, it achieves $1.53\times$ more windows past $99\%$ performance, $2.32\times$ more windows past $98\%$ performance, and $4.11\times$ more windows past $96\%$ performance compared to OPT-6.7B. Additionally, it achieves at least $94\%$ of peak performance $2.75\times$ more often than OPT-6.7B and $11.18\times$ more often than OPT-1.3B. This shows that the policy is able to correctly balance between OPT-6.7B, OPT-1.3B, and OPT-125M, even while faced with an unpredictable workload.

\begin{figure}[h]
\vskip 0.1in
\begin{center}
\centerline{\includegraphics[width=\columnwidth]{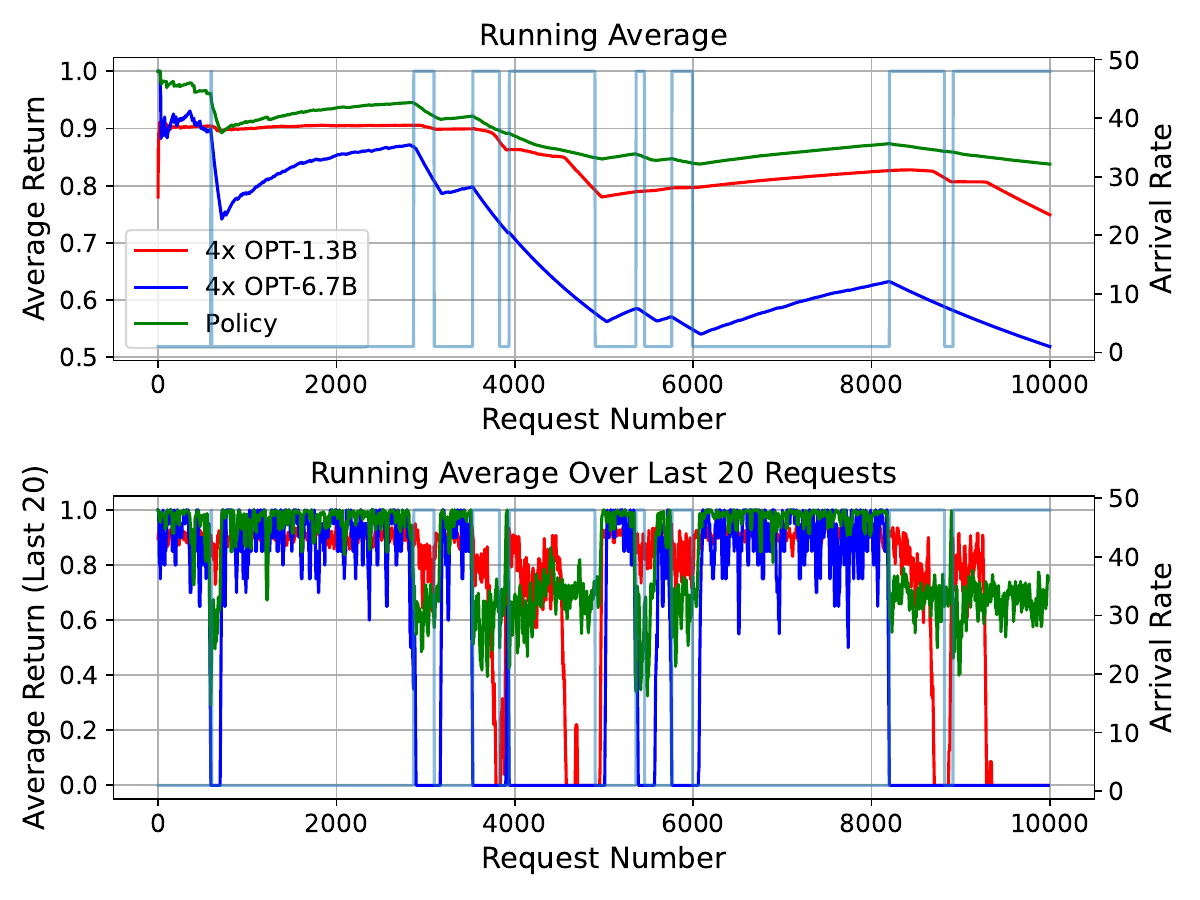}}
\caption{Running total and windowed average over the last 20 requests of performance on the second unpredictable workload. The arrival rate at each step is also shown.}
    \label{fig:env-1-unpredictable-2}
\end{center}
\vskip -0.25in
\end{figure}

We also show performance on the second unpredictable workload. The results are shown in \autoref{fig:env-1-unpredictable-2}. As the results in \autoref{fig:env-1-unpredictable-2} show, the policy outperforms both baselines of OPT-1.3B and OPT-6.7B and is able to adapt to the large changes in arrival rate. Additionally, as shown in \autoref{tab:env-1-unpredictable-2-peak-nums}, the policy has significantly more windows which meet high performance thresholds. It achieves $1.43\times$ more windows past $99\%$ performance, $2.26\times$ more windows past $98\%$ performance, and $3\times$ more windows past $96\%$ performance compared to OPT-6.7B. Additionally, it achieves at least $94\%$ of peak performance $1.68\times$ more often than OPT-6.7B and $28.21\times$ more often than OPT-1.3B. 

\begin{table}[h]
\caption{Number of request windows of size 20 that meet average performance thresholds on the second unpredictable workload.}
\label{tab:env-1-unpredictable-2-peak-nums}
\vskip 0.15in
\begin{center}
\begin{small}
\begin{sc}
\begin{tabular}{lccc}
\toprule
Threshold & Policy & OPT-6.7B & OPT-1.3B \\
\midrule
$= 1.00$       &447&\textbf{1378}&0\\
$\geq 0.99$    &\textbf{1977}&1378&0\\
$\geq 0.98$    &\textbf{3121}&1378&0\\
$\geq 0.96$    &\textbf{4141}&1378&0\\
$\geq 0.94$    &\textbf{4740}&2810&168 \\
\bottomrule
\end{tabular}
\end{sc}
\end{small}
\end{center}
\vskip -0.1in
\end{table}

\subsection{Robustness to Shifts in Task Distribution}
\label{sec:robustness}
When training the policy we assume that tasks are picked uniformly at random in the application's workload.  For some classes of applications, it is possible that the distribution of tasks changes after the policy has been deployed. It is important that the policy still perform well when the task distribution shifts. 

\begin{figure}[h]
\vskip 0.1in
\begin{center}
\centerline{\includegraphics[width=\columnwidth]{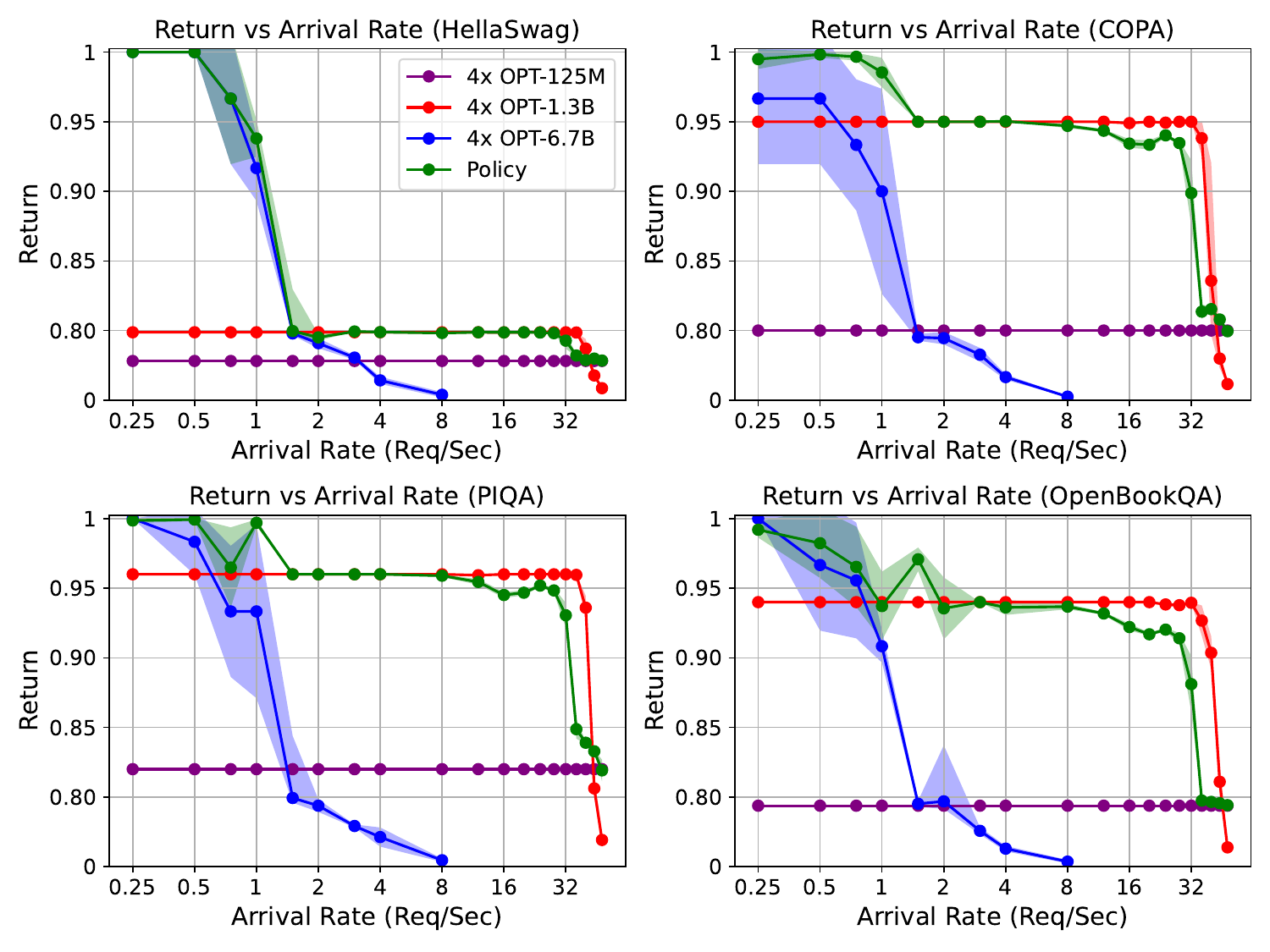}}
\caption{Performance for each of the out-of-distribution workloads, corresponding to sending only one task to the serving system.}
    \label{fig:one-task-1}
\end{center}
\vskip -0.25in
\end{figure}

We show the performance of our policy in the hard deadline setting while drastically changing the task distribution and without performing any additional training iterations. Specifically, we evaluate performance when the serving system is only receiving requests from one class of tasks at a time. The results on the stable workload are shown in \autoref{fig:one-task-1}. Even though the task distribution is drastically different than the training distribution, the policy is still able to achieve better performance than OPT-6.7B OPT-1.3B, and OPT-125M.

\begin{figure}[h]
\vskip 0.1in
\begin{center}
\centerline{\includegraphics[width=\columnwidth]{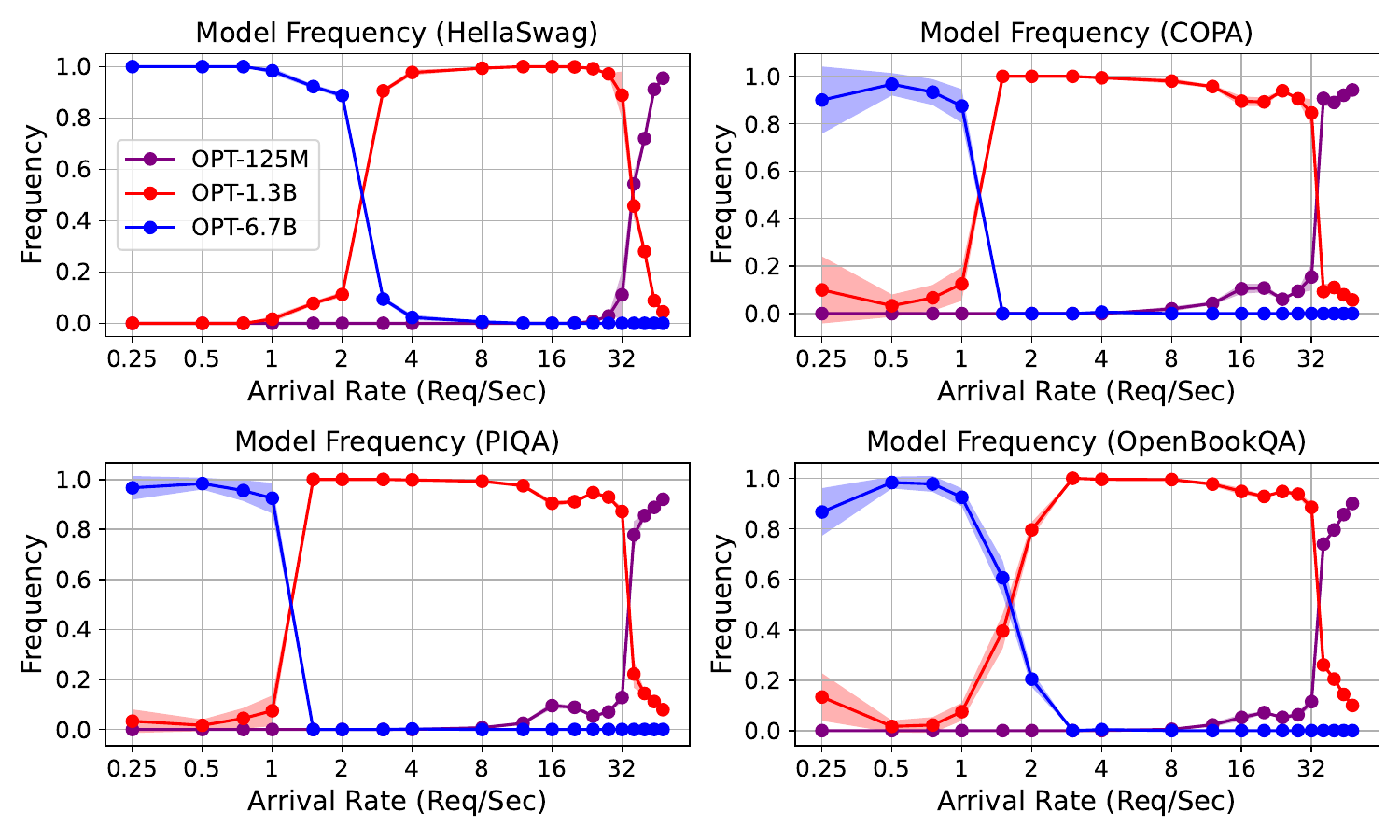}}
\caption{Model selection frequency for each of the out-of-distribution workloads, corresponding to sending only one task to the serving system.}
    \label{fig:one-task-2}
\end{center}
\vskip -0.25in
\end{figure}

We notice that when serving only HellaSwag, the policy closely follows OPT-6.7B's performance at low arrival rates. In contrast, when only serving COPA, PIQA, or OpenBookQA, the policy is able to outperform OPT-6.7B at these low arrival rates. When analyzing the policy's model selection distribution for each of the tasks in \autoref{fig:one-task-2}, we see that this is because the policy favors OPT-6.7B on HellaSwag more than the other tasks. Although this was optimal when tasks were picked uniformly at random, it is slightly less optimal now because the policy may benefit from sending some HellaSwag requests to OPT-1.3B due to randomness in the arrival process. For COPA, PIQA, and OpenBookQA, the policy is able to multiplex better with OPT-6.7B and OPT-1.3B and thus is able to beat OPT-6.7B at lower arrival rates. The out-of-distribution performance outperforms the baselines and highlights the policy's robustness.  

We also evaluate the soft policy's robustness to distribution shift by sending only HellaSwag and COPA with the arrival patterns of the first unpredictable workload using soft deadlines. We detail the results in \autoref{sec:appendix-ood-soft} and show that the soft policy is robust to distribution shift as well.

\subsection{Hardware Utility}
\label{sec:hw-utility}
\begin{figure}[h]
\vskip 0.1in
\begin{center}
\centerline{\includegraphics[width=\columnwidth]{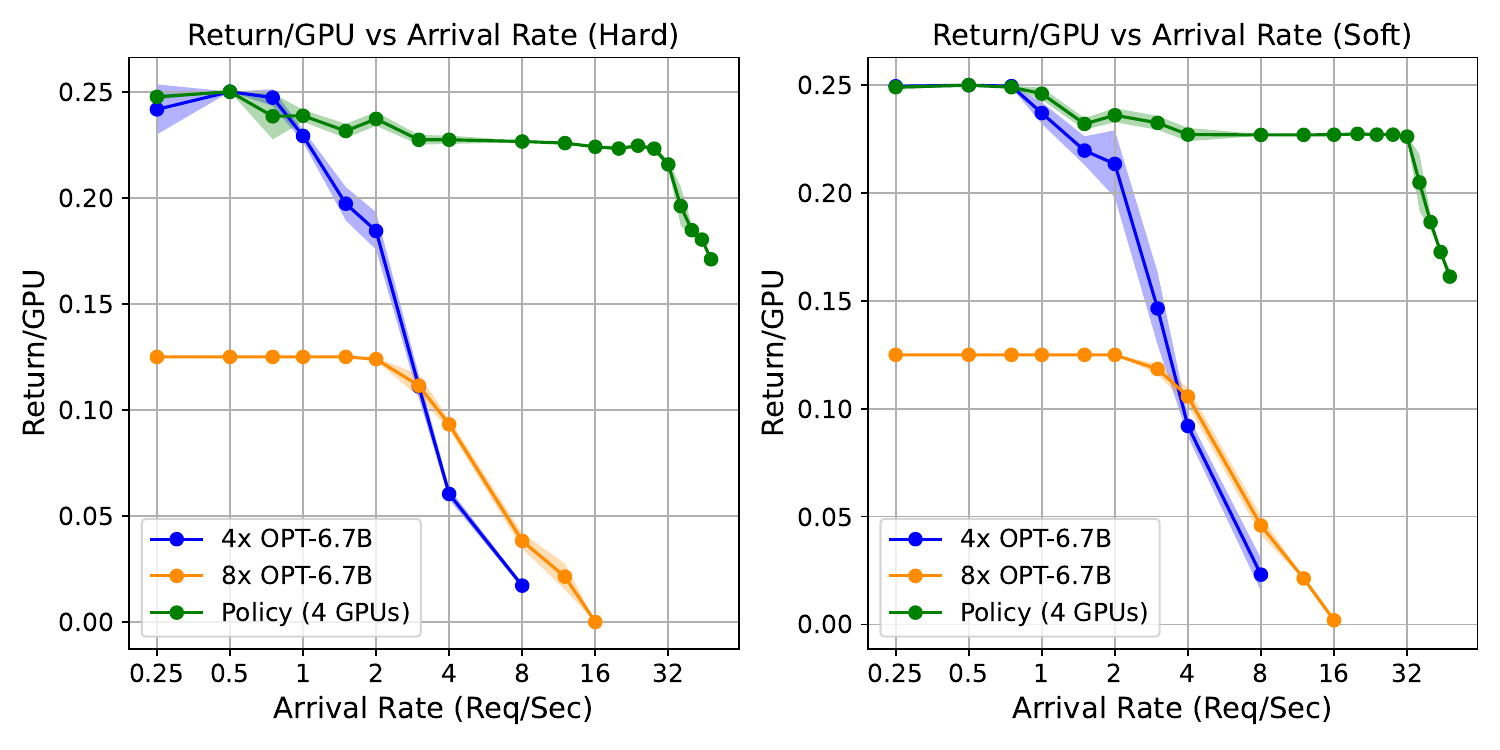}}
\caption{Hardware utility of the policy running on 4 GPUs compared to an OPT-6.7B system running on 8 GPUs. Both hard deadline (left) and soft deadline (right) performance is shown.}
    \label{fig:env-1-hw-utility}
\end{center}
\vskip -0.25in
\end{figure}

\begin{figure}[h]
\vskip 0.1in
\begin{center}
\centerline{\includegraphics[width=\columnwidth]{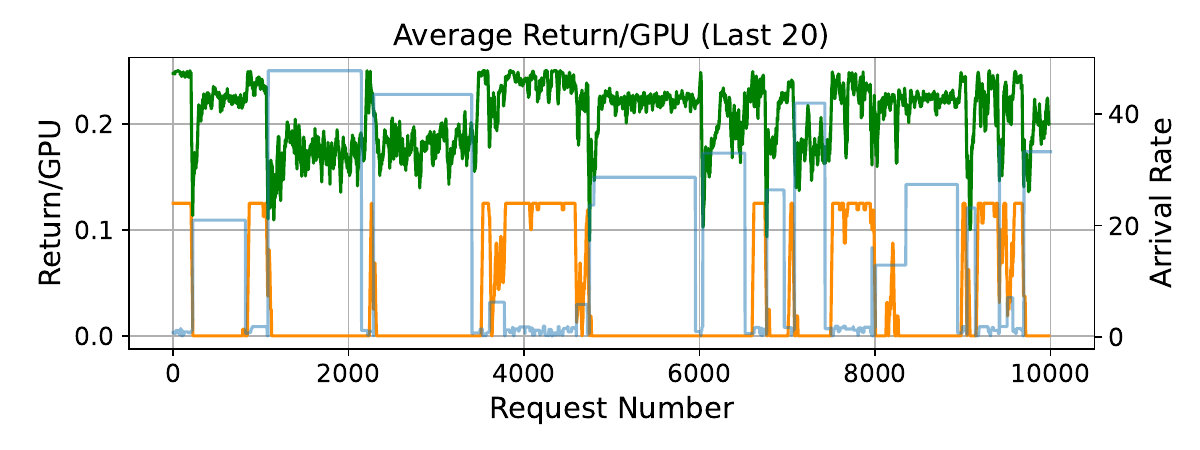}}
\caption{Windowed average over the last 20 requests of the hardware utility of the policy running on 4 GPUs compared to an OPT-6.7B system running on 8 GPUs, on the first unpredictable workload.}
    \label{fig:env-1-hw-utility-unpredictable}
\end{center}
\vskip -0.25in
\end{figure}

Our learned router leads to increased performance per hardware unit, which we call hardware utility. To demonstrate this, we compare the performance of running OPT-6.7B replicated on 8 GPUs against running our policy on 4 GPUs. We show the results in \autoref{fig:env-1-hw-utility} on the stable workload using both hard and soft deadlines. When showing the performance, we normalize by the number of GPUs to capture the hardware utility. As \autoref{fig:env-1-hw-utility} shows, the policy running on 4 GPUs significantly outperforms both the 4 and 8 GPU OPT-6.7B system on a per-GPU basis. Additionally, we compare the policy's hardware utility against the 8 GPU system on the first unpredictable workload and show the results in \autoref{fig:env-1-hw-utility-unpredictable}. The policy achieves higher hardware utility than the 8 GPU system $97.03\%$ of the time. On average, the policy's hardware utility is $3.94\times$ higher than the 8 GPU system. Additionally, the policy serves past $90\%$ of peak performance $1.51\times$ more often than the 8 GPU OPT-6.7B system, when running on just 4 GPUs. Since reserving many GPU instances and having the budget to pay for them are both difficult tasks, these results highlight an important advantage for learned best-effort serving over static serving.

\subsection{Different Deadlines}
\label{sec:diff-deadlines}
\begin{figure}[h]
\vskip 0.1in
\begin{center}
\centerline{\includegraphics[width=\columnwidth]{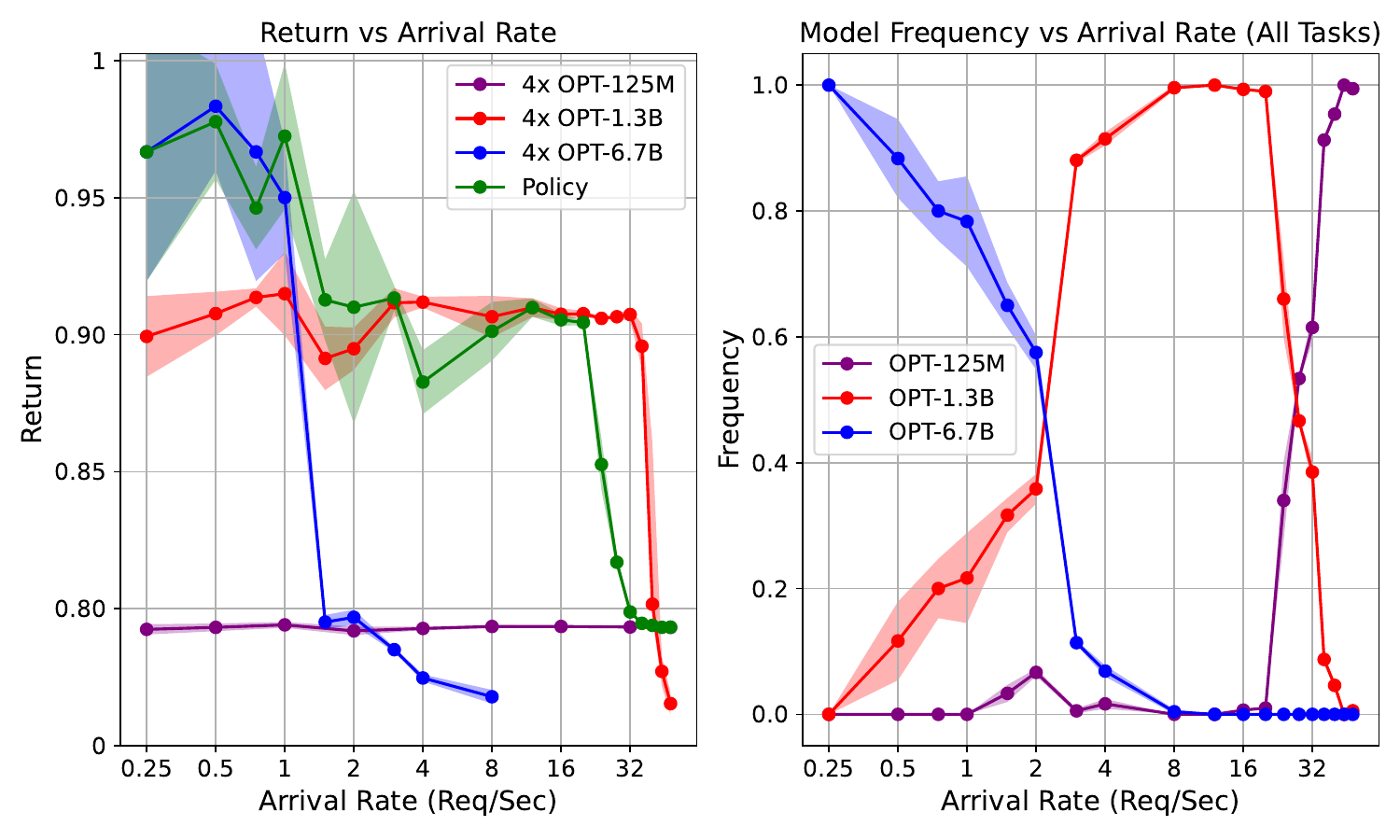}}
\caption{The left figure shows the performance with loose OpenBookQA and tight COPA deadlines. The right figure shows the distribution of model selection from the policy.}
    \label{fig:different-deadlines-1}
\end{center}
\vskip -0.25in
\end{figure}

\begin{figure}[h]
\vskip 0.1in
\begin{center}
\centerline{\includegraphics[width=\columnwidth]{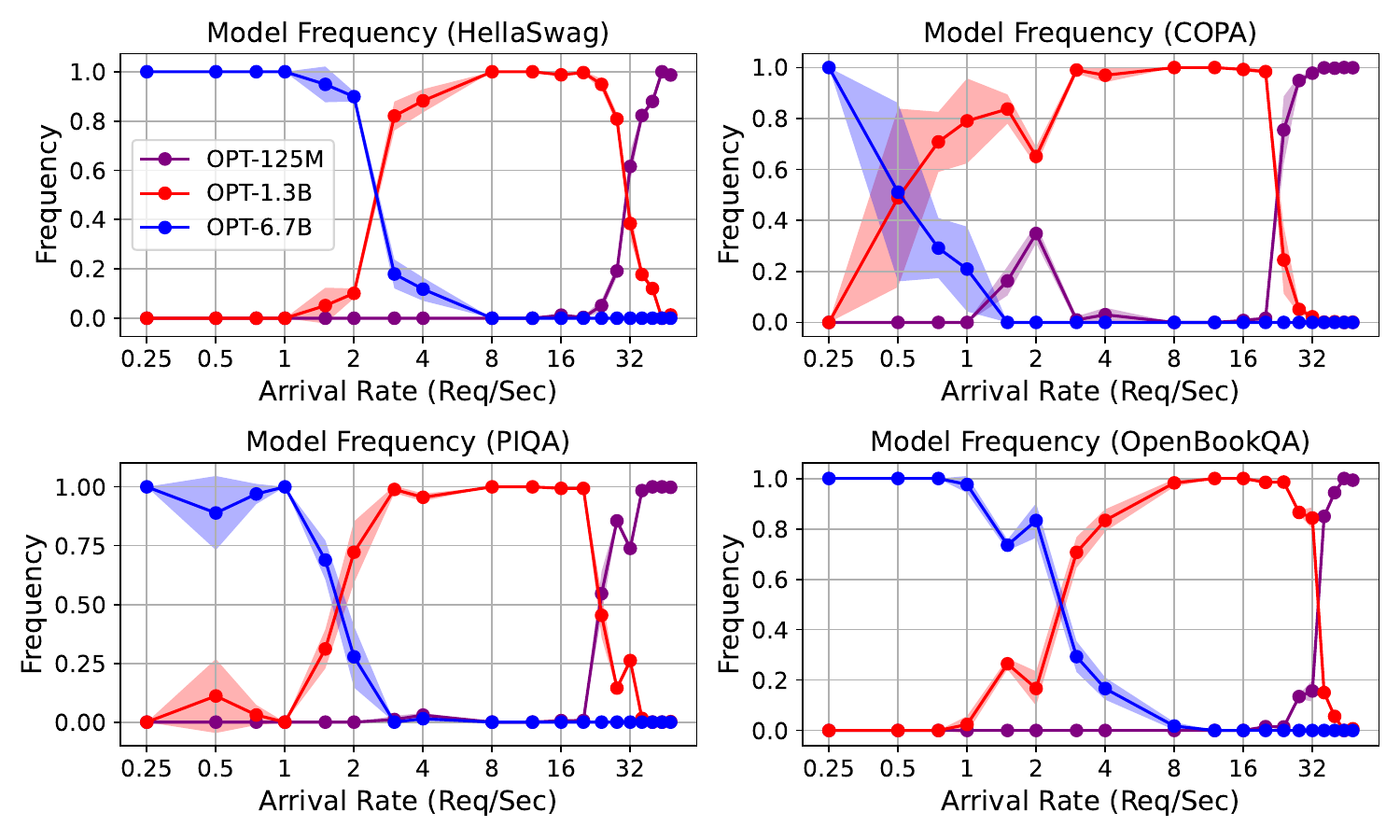}}
\caption{Model selection frequency for each individual task with loose OpenBookQA and tight COPA deadlines.}
    \label{fig:different-deadlines-2}
\end{center}
\vskip -0.25in
\end{figure}

Certain applications may want different deadlines to be associated with different tasks. To see if the policy can handle this setting, we double the deadline for OpenBookQA and tighten the deadline for COPA by 20\%. We train the policy for 800,000 iterations using hard deadlines and show both the performance and model selection in \autoref{fig:different-deadlines-1} and \autoref{fig:different-deadlines-2}. We see that the policy is able to learn with different deadlines and outperforms the static serving baselines as before. Additionally, compared to the task-specific selection distribution in \autoref{fig:env-1-hard-task-dist}, the policy sends significantly more OpenBookQA requests to OPT-6.7B, as the deadline is now looser. Compared to the other tasks in the system, the policy waits for a higher arrival rate before it switches sending OpenBookQA requests to OPT-125M instead of OPT-1.3B. Similarly, the policy significantly reduces OPT-6.7B usage for COPA in favor of OPT-1.3B due to the tighter deadlines.

\section{Discussion}
Best-effort serving with dynamic routing is an efficient approach for developers looking to scale their latency-sensitive applications. It is also a good option for applications whose client request rates have wide fluctuations, which is common in many practical settings~\citep{zhang2021faster}. Viewed through another lens, learned best-effort serving allows higher quality during low system load than statically serving a smaller model to handle periods of high loads. There are also a wide range of system environments in which best-effort serving is applicable. By formulating model serving as a reinforcement learning problem, application developers have the flexibility to adapt the reward function in order to meet their application requirements. For example, they may choose to up-weight the reward on prioritized requests coming from paid users. Lastly, certain applications may want a latency deadline on the first generated token, which can easily be incorporated into the reward function as well.

\section{Conclusion}
Rather than serving LLMs at a fixed model size, we propose a best-effort serving paradigm with a learned router that maximizes holistic performance, which jointly captures quality and latency. We train our router using deep reinforcement learning methods with minimal hyper-parameter tuning and outperform static serving baselines in a variety of workloads. We show that the router is robust to changes in both the arrival patterns and task distribution. Additionally, learned best-effort serving allows for significantly higher hardware utility compared to static serving. We imagine best-effort serving with dynamic routing to be a cheap and efficient paradigm for latency-sensitive applications.

\bibliography{paper}
\bibliographystyle{icml2024}

\newpage
\appendix
\onecolumn
\section{Training Details}
\label{sec:appendix-training-details}
We use a discount rate of 0.99, a learning rate of 0.0001, and a batch size of 1024. In our Double Q-learning implementation, the target network is updated every 500 iterations. For exploration, we use an epsilon-greedy strategy. We use an NVIDIA TITAN RTX for GPUs and an Intel Xeon Gold 6126 processor as our CPU.

\section{Workload Details}
\label{sec:appendix-workload-details}
There are three workloads we use - 1 stable and 2 unpredictable. The first represents a stable workload in which client requests arrive in the system as a Poisson process with a fixed rate for a set period of time. The second and third workload represent unpredictable workloads in which the arrival rate of requests rapidly switches due to an underlying stochastic process that controls the arrival rate and its duration. The difference between the second and third workload is that the second workload assumes that the system spends the same amount of time (in expectation) in each arrival rate, while the third workload assumes that the system serves the same amount of requests (in expectation) in each arrival rate before switching. On the unpredictable workloads, we estimate the arrival rate by taking a running average of the last 5 arrivals. It is also possible for application developers to use prior information on the arrival patterns, but we do not use any. Since a near-zero variance estimate of the arrival rate may be obtained on the stable workloads, we give the agent the true arrival rate.

For the first unpredictable workload, we randomly vary the arrival rate and the number of requests served at that arrival rate before switching to the next arrival rate. With 90\% probability, we randomly pick an arrival rate between 0.25 and 2. With 8\% probability, we randomly pick an arrival rate between 2 and 40. With the remaining 2\% probability, we randomly pick an arrival rate between 40 and 48. The number of requests served at the arrival rate is a geometric random variable with mean 20 $\times$ arrival rate. There are 10,000 requests in total.

In the second unpredictable workload, the arrival rate randomly fluctuates between 1 request per second and 48 requests per second. The first unpredictable workload assumed that the expected time spent in each arrival rate is the same. With the new workload, we assume that the number of expected requests seen in each arrival rate is the same, meaning large arrival rates occupy less real time in the system than small arrival rates. We use a geometric random variable with mean 500 to determine the number of requests to serve at an arrival rate before picking the next arrival rate. There are 10,000 requests in total.

\section{Additional Task Distribution Shift Experiment}
\label{sec:appendix-ood-soft}
\begin{figure}[h]
\vskip 0.1in
\begin{center}
\centerline{\includegraphics[width=150pt]{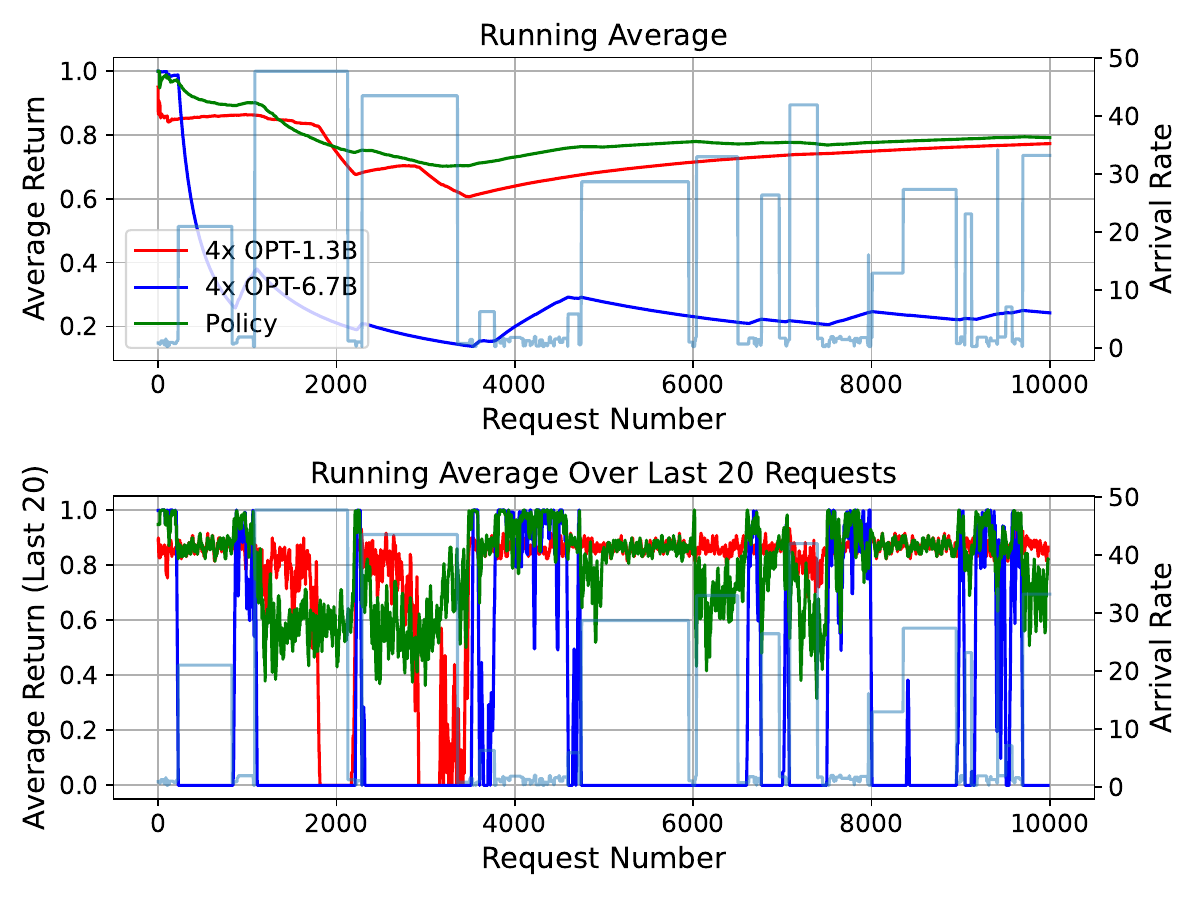}}
\caption{Running total and windowed average over the last 20 requests of performance on the first unpredictable workload while only serving HellaSwag and COPA with soft deadlines. The arrival rate at each step is also shown.}
    \label{fig:ood-task-unpredictable}
\end{center}
\vskip -0.25in
\end{figure}

We also evaluate the soft policy's resistance to shifts in the task distribution on an unpredictable workload. The workload is the first unpredictable workload in \autoref{sec:env-1-unpredictable} but requests are chosen uniformly at random between only HellaSwag and COPA. As shown in \autoref{fig:ood-task-unpredictable}, the policy is still able to outperform the baselines.

\end{document}